\documentclass[11pt,letterpaper]{article}
\usepackage{naaclhlt2016}
\usepackage{times}
\usepackage{latexsym}
\usepackage{graphicx}
\usepackage{amssymb}
\usepackage{url}
\usepackage{multirow}

\naaclfinalcopy 
\def\naaclpaperid{***} 

\title{Text Readability Assessment for Second Language Learners}

\author{Menglin Xia \and Ekaterina Kochmar \and Ted Briscoe\\
	    The ALTA Institute\\
	    Computer Laboratory\\
	    University of Cambridge\\
	    Cambridge, CB3 0FD, UK\\
	    \{{\tt mx223,ek358,ejb1}\}{\tt @cl.cam.ac.uk}
}

\date{}

\begin{document}

\maketitle

\begin{abstract}

This paper addresses the task of readability assessment for the texts aimed at second language (L2) learners. One of the major challenges in this task is the lack of significantly sized level-annotated data. For the present work, we collected a dataset of CEFR-graded texts tailored for learners of English as an L2 and investigated text readability assessment for both native and L2 learners. We applied a generalization method to adapt models trained on larger native corpora to estimate text readability for learners, and explored domain adaptation and self-learning techniques to make use of the native data to improve system performance on the limited L2 data. In our experiments, the best performing model for readability on learner texts achieves an accuracy of $0.797$ and $PCC$ of $0.938$. 

\end{abstract}

\section{Introduction}

Developing reading ability is an essential part of language acquisition. However, finding proper reading materials for training language learners at a specific level of proficiency is a demanding and time-consuming task for English instructors as well as the readers themselves. To automate the process of reading material selection and the assessment of reading ability for non-native learners, a system that focuses on text readability analysis for L2 learners can be developed. Such a system enhances many pedagogical applications by supporting readers in their second language education. 

Text readability, which has been formally defined as the sum of all elements in textual material that affect a reader's understanding, reading speed, and level of interest in the material \cite{dale1949concept}, is influenced by multiple variables. These may include the style of writing, its format and organization, reader's background and interest as well as various contextual dimensions of the text, such as its lexical and syntactic complexity, level of conceptual familiarity, logical sophistication and so on. 

The choice of the criteria to measure readability often depends upon the need and characteristics of the target readers. Most of the studies so far have evaluated text difficulty as judged by native speakers, despite the fact that text comprehensibility can be perceived very differently by L2 learners. In the case of L2 learners, due to the difference in the pace of language acquisition, the focus in readability measures often differs from that for native readers. For example, the grammatical aspects of readability usually contribute more to text comprehensibility for L2 learners than the conceptual cognition difficulty of the reading material \cite{callan2007combining}. A system that is tailored towards learner's perception of reading difficulty can produce more accurate estimation of text reading difficulty for non-native readers and thus better facilitate language learning. 

One of the major challenges for a data-driven approach to text readability assessment for L2 learners is that there is not enough significantly sized, properly annotated data for this task. At the same time, text readability assessment in general has been previously studied by many researchers and there are a number of existing corpora aimed at native speakers that can be used. To address the problem, we compiled a collection of texts that are tailored for L2 learners' readability and looked at several approaches to make use of existing native data to estimate readability for L2 learners. 

In sum, the contribution of our work is threefold. First, we develop a system that produces state-of-the-art estimation of text readability, exploit a range of readability measures and investigate their predictive power. Second, we focus on readability for L2 learners of English and present a level-graded dataset for non-native readability analysis. Third, we explore methods that help to make use of the existing native corpora to produce better estimation of readability when there is not enough data aimed at L2 learners. Specifically, we apply a generalization method to adapt models trained on native data to estimate text readability for learners, and explore domain adaptation and self-training techniques to improve system performance on the data aimed at L2 learners. To the best of our knowledge, these approaches have not been applied in readability experiments before. The best performing model in our experiments achieves an accuracy ($ACC$) of $0.797$ and Pearson correlation coefficient ($PCC$) of $0.938$.

\section{Related Work}

        \subsection{Automated Readability Assessment}

       Many previous studies on text readability assessment have used machine learning based approaches, which enable investigation of a broader set of linguistic features. Si and Callan \shortcite{si2001statistical} and Collins-Thompson and Callan \shortcite{collins2004language} were among the early works on statistical readability assessment. They applied unigram language models and na{\"i}ve Bayes classification to estimate the grade level of a given text. Experiments showed that the language modelling approach yields better results in terms of accuracy than the traditional readability formulae, such as the the Flesch-Kincaid score \cite{kincaid1975derivation}. Schwarm and Ostendorf \shortcite{schwarm2005reading} extended this method to multiple language models. They combined traditional reading metrics with statistical language models as well as some basic parse tree features and then applied an SVM classifier. Heilman et al. \shortcite{callan2007combining,heilman2008analysis} expanded the feature set to include certain lexical and grammatical features extracted from parse trees while using a linear regression model to predict the grade level. 
        
        Pitler and Nenkova \shortcite{pitler2008revisiting} and Feng et al. \shortcite{feng2010comparison} were the first to introduce discourse-based features into the framework. The experiments with discourse features demonstrated promising results in predicting the readability level of text for both classification and regression approaches.
        
		Kate et al. \shortcite{kate2010learning} looked at both the effect of the feature choice and the machine learning framework choice on performance, and found that the improvement resulting from changing the framework is smaller than that from changing the features. 

		\subsection{Readability Assessment for L2 Learners}
		
		Most previous work on readability assessment is directed at predicting reading difficulty for native readers. Several efforts in developing automated readability assessment that take L2 learners into consideration have emerged since 2007. Heilman et al. \shortcite{callan2007combining} tested the effect of grammatical features for both L1 (first language) and L2 readers and found that grammatical features play a more important role in L2 readability prediction than in L1 readability prediction. Vajjala and Meurers \shortcite{vajjala2012improving} combined measures from Second Language Acquisition research with traditional readability features and showed that the use of lexical and syntactic features for measuring language development of L2 learners has a substantial positive impact on readability classification. They observed that lexical features perform better than syntactic features, and that the traditional features have a good predictive power when used with other features. Shen et al. \shortcite{shen2013language} developed a language-independent approach to automatic text difficulty assessment for L2 learners. They treated the task of reading level assessment as a discriminative problem and applied a regression approach using a set of features that they claim to be language-independent. However, most of these studies have used textual data annotated with the readability levels for native speakers of English rather than L2 learners specifically.
		
		While the majority of work on automated readability assessment are for English, studies on L2 readability in other languages, including French \cite{franccois2012ai}, Portuguese \cite{branco2014assessing}, and Swedish \cite{volodina2015}, are also emerging. These studies generally use textbook materials with readability levels assigned by publishers or language instructors.
		
		Overall, study of automatic readability analysis for L2 learners is still in its early stages, mainly due to the lack of available well-labelled data annotated with the readability levels for L2 learners.

\section{Data}

\subsection{Native Data: the WeeBit Corpus}
\label{nativedata}
        Among the existing publicly available corpora, the WeeBit corpus created by Vajjala and Meurers \shortcite{vajjala2012improving} is one of the largest datasets for readability analysis. The WeeBit corpus is composed of articles targeted at readers of different age groups from two sources, the Weekly Reader magazine and the BBC-Bitesize website. Within the dataset, the Weekly Reader data consists of texts covering age-appropriate non-fictional content for four grade levels, corresponding to children of ages between 7-8, 8-9, 9-10 and 10-12 years old. The BBC-Bitesize website data is targeted at two grade levels, for ages between 11-14 and 14-16. The two datasets are merged to form the WeeBit corpus, with the targeted ages used to assign readability levels.
		
		A copy of the original WeeBit corpus was obtained from the authors \cite{vajjala2012improving}. The texts are webpage documents stored in raw HTML format. We have identified that some texts contain broken sentences or extraneous content from the webpages, such as copyright declaration and links, that correlate with the target labels in a way which is likely to artificially boost performance on the task and would not generalize well to other datasets. To avoid that, we re-extracted texts from the raw HTML and discarded text documents that do not contain proper reading passages. Table \ref{weebit} shows the distribution of texts in the modified dataset.

	    \begin{table}[t]
			\centering
			\small
			\resizebox{\columnwidth}{!}{\begin{tabular}{l|c|c|c|c|c}
				\hline
				& Level1 & Level2 & Level3 & Level4 & Level5  \\ \hline \hline
				age group                   & 7-8    & 8-9    & 9-10   & 10-14  & 14-16 \\ \hline
				original corpus & 629    & 801    & 814    & 1969   & 3500  \\ \hline
				modified corpus & 529    & 767    & 801    & 1288   & 845   \\ \hline
			\end{tabular}}
		    \caption{Number of documents in the original and modified WeeBit corpus}
			\label{weebit}

		\end{table}

\subsection{L2 Data: the Cambridge Exams dataset}
    Most work on readability assessment has been done on native corpora with age-specific reading levels \cite{schwarm2005reading,feng2010comparison}. Such texts are aimed not at L2 learners but rather at native-speaking children of different ages. Therefore, the level annotation in such texts is arrived at using criteria different from those that are relevant for L2 readers. The lack of significantly sized L2 level-annotated data raises a problem for readability analysis aimed at L2 readers. To tackle this, we created a dataset with texts tailored for L2 learners' readability specifically. 
     
    We have collected a dataset composed of reading passages from the five main suite Cambridge English Exams (KET, PET, FCE, CAE, CPE).\footnote{\url{http://www.cambridgeenglish.org}} These five exams are targeted at learners at A2--C2 levels of the Common European Framework of Reference (CEFR) \cite{cefr2001common}.\footnote{The CEFR determines foreign language proficiency at six levels in increasing order: A1 and A2, B1 and B2, C1 and C2.}  The documents are harvested from all the tasks in the past reading papers for each of the exams. The Cambridge English Exams are designed for L2 learners specifically and the A2--C2 levels assigned to each reading paper can be treated as the level of reading difficulty of the documents for the L2 learners.\footnote{We are aware that the type of the task may also have an effect on the reading difficulty of the texts, but this is ignored at this stage.} Table \ref{cameng} shows the number of documents at each CEFR level across the dataset. The data will be available at \url{http://www.cl.cam.ac.uk/~ mx223/cedata.html}.
    
      Experimenting on the language testing data annotated with the L2 learner readability levels is one of the contributions of this research. Most previous work on readability assessment for English have relied on the data annotated with readability levels aimed at native speakers. In this work, we use language testing data with the levels assigned based on L2 learner levels, and we believe that this level annotation is more appropriate for text readability assessment for L2 learners than using texts with the level annotation aimed at native speakers.
    
	\begin{table}[t]
		\centering
		\small
		\resizebox{\columnwidth}{!}{\begin{tabular}{l|c|c|c|c|c}
			\hline
			Exams	&	KET	&	PET	&	FCE	&	CAE	&	CPE \\ 
			\hline
			\hline
			targeted level	&	A2	&	B1	&	B2	&	C1	&	C2 \\ \hline
			\# of docs	&	64	&	60	&	71	&	67	&	69 \\ \hline
			avg. len. of text & 14.75 & 19.48 & 38.07 & 45.76 & 39.97 \\ \hline
		\end{tabular}}
		\caption{Statistics for the Cambridge English Exams data}\label{cameng}
	\end{table}

\section{Readability Measures}
    This section describes the range of linguistic features explored and the machine learning framework applied to the WeeBit data that constitute a general readability assessment system. The set of features used in our experiments is an extension to those used in previous work \cite{feng2010comparison,pitler2008revisiting,vajjala2012improving,vajjala2014readability}, and their predictive power for reading difficulty assessment is investigated in our experiments. We have extended the feature set with the EVP-based features, GR-based complexity measures and the combination of language modeling features that have not been applied to readability assessment before.
    
\subsection{Features}    
		
{\bf Traditional Features}
The traditional features are easy-to-compute representations of superficial aspects of text. The metrics that are considered include: the number of sentences per text, average and maximum number of words per sentence, average number of characters per word, and average number of syllables per word. Two popular readability formulas are also included: the Flesch-Kincaid score \cite{kincaid1975derivation} and the Coleman-Liau readability formula \cite{coleman1975computer}.

{\bf Lexico-semantic Features}
Vocabulary knowledge is one of the most important aspects of reading comprehension \cite{collinsthompson2014survey}. Lexico-semantic features provide information about the difficulty or familiarity of vocabulary in the text.

A widely used lexical measure is the \textit{type-token ratio} (TTR), which is the ratio of the number of unique word tokens (referred to as types) to the total number of word tokens in a text. However, the conventional TTR is influenced by the length of the text. \textit{Root TTR} and \textit{Corrected TTR}, which take the logarithm and square root of the text length instead of the direct word count as denominator, can produce a more unbiased representation and are included in the experiment.

Part of speech (POS) based lexical variation and lexical density measures \cite{lu2011carpus} are also examined. \textit{Lexical variation} is defined as the type-token ratio of lexical items such as nouns, adjectives, verbs, adverbs and prepositions. \textit{Lexical density} is defined as the proportion of the five classes of lexical items in all word tokens. The percentage of content words (nouns, verbs, adjectives and adverbs) and function words (all the remaining POS types) are two other indicators of lexical density. 
				
Vajjala and Meurers \shortcite{vajjala2012improving,vajjala2014readability} reported in their readability classification experiment that the proportion of words in the text that are found in the Academic Word List is one of the most predictive measures among all the lexical features they considered. The Academic Word List \cite{coxhead2000new} is comprised of words that frequently occur across all topic ranges in an academic text corpus. The {\em proportion of academic vocabulary words} in the text can be viewed as another measure of lexical complexity.
				
A similar but more refined approach to estimate lexical complexity is based on the use of the {\em English Vocabulary Profile} (EVP).\footnote{\url{http://www.englishprofile.org/}} The EVP is an online vocabulary resource that contains information about which words and phrases are acquired by learners at each CEFR level. It is collected from the Cambridge Learner Corpus (CLC), a collection of examination scripts written by learners from all over the world \cite{capel2012completing}. It provides a more fine-grained lexical complexity measure that captures the relative difficulty of each word by assigning the word difficulty to one of the six CEFR levels. Additionally, the EVP indicates the word difficulty for L2 learners rather than native speakers, which makes it more informative in non-native readability analysis. In our experiments, the proportion of words at each CEFR level is calculated and added to the feature set. 

{\bf Parse Tree Syntactic Features} 
A number of {\em syntactic measures} based on the RASP parser output \cite{briscoe2006second} are used to describe the grammatical complexity of text, including average parse tree depth, and average number of noun, verb, adjective, adverb, prepositional phrases and clauses per sentence. 

{\em Grammatical relations} (GR) between constituents in a sentence may also affect the judgement of syntactic difficulty. Yannakoudakis \shortcite{yannakoudakis2013automated} applied 24 GR-based complexity measures in essay scoring and showed good results. These complexity measures capture the grammatical sophistication of the text through the representation of the distance between the sentence constituents. For instance, these measures calculate the longest/average distance in the GR sets generated by the parser and the average/maximum number of GRs per sentence. A set of 24 GR-based measures used by Yannakoudakis \shortcite{yannakoudakis2013automated} are generated by RASP for each sentence. We take the average of these measures across the text to incorporate the GR-related aspect of its syntactic difficulty.

Other types of complexity measures that are derived from the parser output include: {\em cost metric}, which is the total number of parsing actions performed for generating the parse tree; {\em ambiguity of the parse}, and so on. A total number of 114 non-GR based complexity measures are extracted. These complexity measures are averaged across the text and used to model finer details of the syntactic difficulty of the text.

{\bf Language Modeling Features}
Statistical language modeling (LM) provides information about distribution of word usage in the text and is in fact another way to describe the lexical dimension of readability. To avoid over-fitting to the WeeBit data, two types of language modeling based features are extracted using the SRILM toolkit \cite{stolcke2002srilm}: (1) {\em word token n-gram models}, with $n$ ranging from $1$ to $5$, trained on the British National Corpus (BNC), and (2) {\em POS n-grams}, with $n$ ranging from $1$ to $5$, trained on the five levels in the WeeBit corpus itself. The LMs are used to score the text with log-likelihood and perplexity. 

{\bf Discourse-based Features}
	Discourse features measure the cohesion and coherence of the text. Three types of discourse-based features are used.
				
	{\bf (1) Entity density features}
	
	Previous work by Feng et al. \shortcite{feng2009cognitively,feng2010comparison} has shown that entity density is strongly associated with text comprehension.  An entity set is a union of named entities and general nouns (including nouns and proper nouns) contained in a text, with overlapping general nouns removed. Based on this, {\em 9 entity density features}, including the total number of all/unique entities per document, the average number of all/unique entities per sentence, percentage of named entities per sentence/document, percentage of named entities in all entities, percentage of overlapping nouns removed, and percentage of unique named entities in all unique entities, are calculated.

	{\bf (2) Lexical chain features}
	
	Lexical chains model the semantic relations among entities throughout the text. The lexical chaining algorithm developed by Galley and McKeown \shortcite{galley2003improving} is implemented. The semantically related words for the nouns in the text, including synonyms, hypernyms, and hyponyms, are extracted from the WordNet \cite{miller1995wordnet}. Then for each pair of the nouns in the text, we check whether they are semantically related. Finally, lexical chains are built by linking semantically related nouns in text. A set of {\em 7 lexical chain-based} features are computed, including total number of lexical chains per document, total number of lexical chains normalized with text length, average/maximum lexical chain length, average/maximum lexical chain span, and the number of lexical chains that span more than half of the document.\footnote{The {\em length} of a chain is the number of entities contained in the chain. The {\em span} of a chain is the distance between the indexes of the first and the last entities in the chain.}

	{\bf (3) Entity grid features}
	
	Another entity-based approach to measure text coherence is the entity grid model introduced by Barzilay and Lapata \shortcite{barzilay2008modeling}. They represented each text by an entity grid, which is a two-dimensional array that captures the distribution of discourse entities across text sentences. Each grid cell contains the grammatical role of a particular entity in the specified sentence: whether it is a subject (S), object (O), neither a subject nor an object (X), or absent from the sentence (-). A local entity transition is defined as the transition of the grammatical role of an entity from one sentence to the following sentence. In our experiments, we used the Brown Coreference Toolkit v1.0 \cite{eisner2011extending} to generate the entity grid for the documents. The {\em probabilities of the 16 types} of local entity transition patterns are calculated to represent the coherence of the text.

\subsection{Implementation and Evaluation}

     In our experiments, we cast readability assessment as a supervised machine learning problem. In particular, a pairwise ranking approach is adopted and compared with a classification method. We believe that the reading difficulty of text is a continuous rather than discrete variable. Text difficulty within a level can also vary. Instead of assigning an absolute level to the text, treating readability assessment as a ranking problem allows prediction of the relative difficulty of pairs of documents, which captures the gradual nature of readability better. Because of this, we hypothesize that the ranking model can generalize better to unseen texts and texts with different level annotation.

    Support vector machines (SVM) have been used in the past for readability assessment by many researchers and have consistently yielded better results when compared to other statistical models for the task \cite{kate2010learning}. We use the LIBSVM toolkit \cite{chang2011libsvm} to implement both multi-class classification and pairwise ranking. Five-fold cross validation is used for evaluation. We report two popular performance metrics, accuracy ($ACC$) and Pearson correlation coefficient ($PCC$), and use pairwise accuracy to evaluate ranking models. Pairwise accuracy is defined as the percentage of instance pairs that the model ranked correctly. It should be noted that accuracy and pairwise accuracy are not directly comparable. Thus, $PCC$ is introduced to compare the results of the classification and the ranking models.

\subsection{Results}

			\begin{table}[t]
				\centering
				\small
				\resizebox{\columnwidth}{!}{
					\begin{tabular}{l|c|c||c|c}
					\hline
					\multicolumn{1}{l|}{\multirow{2}{*}{feature set}}& \multicolumn{2}{c||}{Classification} & \multicolumn{2}{c}{Ranking} \\
					\cline{2-5}
					 & ACC & PCC & pairwise ACC & PCC	\\
					\hline
					traditional & 0.586 & 0.770 & 0.862 & 0.704\\
					\hline
					lexical & 0.578 & 0.726 & 0.863 & 0.743\\
					\hline
					syntactic & 0.599 & 0.731 & 0.824 & 0.692\\
					\hline
					LM & 0.714 & 0.848 & 0.872 & 0.769  \\
					\hline
					discourse & 0.563 & 0.688 & 0.848 & 0.659\\
					\hline
					all combined & \textbf{0.803} & \textbf{0.900} & \textbf{0.924} & \textbf{0.848}\\
					\hline
			    \end{tabular}
			    }			
				\caption{Classification and ranking results on the WeeBit corpus with feature sets grouped by their type}
			    \label{individual_result}
			\end{table}
			
	In predicting the text reading difficulty on the WeeBit data, the best result is achieved with a combination of all features and a classification model, with $ACC$$=$$0.803$ and $PCC$$=$$0.900$. We performed ablation tests and found that all feature sets have contributed to the overall model performance. Although there have been readability assessment studies on similar datasets, the results obtained in our experiments are not directly comparable to those. One of the major reasons is the modifications that we have made to the corpus (as discussed in Section \ref{nativedata}). Vajjala and Meurers \shortcite{vajjala2012improving} reported that a multilayer perceptron classifier using three traditional metrics alone yielded an accuracy of $70.3\%$ on their version of the WeeBit corpus. Their final system achieved a classification accuracy of $93.3\%$ on the five-class corpus. Nonetheless, the best system in our experiments yields results competitive to most existing studies. For reference, Feng et al. \shortcite{feng2010comparison} reported an accuracy of $74.01\%$ using a combination of discourse, lexical and syntactic features for readability classification on their Weekly Reader Corpus and an accuracy of $63.18\%$ when using all feature sets described in Schwarm et al. \shortcite{schwarm2005reading}.
	
	Comparing the classification and the ranking models, we note that the results of the two models vary across feature sets and none of the two models is consistently better than the other. When all features are combined, the classification model outperforms the ranking one. It suggests that a ranking model is not necessarily the best model in predicting readability overall when trained and tested on the same dataset.

\section{Readability Assessment on L2 Data}

	So far we have studied the effect of various readability measures on the task of readability assessment and built two different types of models to predict text difficulty. However, the WeeBit corpus consists of texts aimed at native speakers of different ages rather than at L2 readers. Although there are certain similarities concerning reading comprehension between these two groups, the perceived difficulty of texts can be very different due to the difference in the pace and stages of language acquisition. Since the goal of our research is to automatically detect readability levels for language learners, it would be more helpful to work with data that are directly annotated with reading difficulty for L2 learners.
	
	Ideally, it would be good to train a model directly on text annotated with L2 levels and then use this model to estimate readability for the new texts. However, the Cambridge Exams data we have compiled is relatively  small, and the model trained on it will likely not generalize well. Therefore, we examined several approaches to make use of the WeeBit corpus for readability assessment on the L2 data.

\subsection{Generalization Experiment}

	First, we tested the generalization ability of the classification and ranking models trained on the WeeBit corpus on the Cambridge Exams data to see if it is possible to directly apply the models trained on native data to L2 data. Table \ref{resultgen} reports the results.
	
    \begin{table}[t]
    \centering
    \small
    \resizebox{0.8\columnwidth}{!}{\begin{tabular}{l|c|c||c|c}
    \hline
    & \multicolumn{2}{c||}{classification} & \multicolumn{2}{c}{ranking}                                   \\ \hline
    & ACC              & PCC              & \begin{tabular}[c]{@{}c@{}}pairwise\\ ACC\end{tabular} & PCC   \\ \hline
    \multicolumn{1}{l|}{native data}          & 0.803            & 0.900            & 0.924                                                  & 0.848 \\ \hline
    \multicolumn{1}{l|}{L2 data} & 0.233            & 0.730            & 0.913                                                  & 0.880 \\ \hline
    \end{tabular}}
    \caption{Generalization results of the classification and ranking models trained on native data applied to language testing data }
    \label{resultgen}

    \end{table}
    
    In the case of the multi-class classification model, the accuracy dropped greatly when the model is applied to the L2 dataset, while the correlation remained relatively high. Looking at the confusion matrix of the classifier's predictions on the L2 data (see Table \ref{cm}), we notice that most of the documents in the L2 data are classified into the higher levels of WeeBit by the model. This is because, on average, the Cambridge Exams texts are more difficult than the WeeBit corpus ones which are generally targeted at children of young ages. Thus, the mismatch between the targeted levels has led to poor generalization of the classification model.

    	\begin{table}[t]
			\centering
			\small
			\resizebox{0.5\columnwidth}{!}{
			\begin{tabular}{l|ccccc}
				Levels & 1 & 2 & 3 & 4 & 5 \\
				\hline
				A2 & 4 & 0 & 55 & 4 & 1 \\
				B1 & 0 & 0 & 24 & 6 & 30 \\
				B2 & 0 & 1 & 1 & 4 & 65 \\
				C1 & 0 & 0 & 0 & 3 & 64\\
				C2 & 0 & 0 & 0 & 0 & 69 \\
			\end{tabular}}
			\caption{Confusion matrix of the classification model on the language testing data}
			\label{cm}

		\end{table}

    In contrast, for the ranking model, both evaluation measures are relatively unharmed when the model is applied to the L2 data. It shows that, when generalizing to an unseen dataset, the estimation produced by the ranking model is able to maintain a high pairwise accuracy and correlation with the ground truth. We believe that this is because the ranking model does not try to band the documents into one of the levels on a different basis of difficulty annotation. Instead, pairwise ranking captures the relative reading difficulty of the documents, and therefore the resulting ranked positions of the documents are closer to the ground truth compared to the classification model.

\subsection{Mapping Ranking Scores to CEFR Levels}
\label{sec:mapping}

		From the generalization experiment we can conclude that ranking is more accurate in predicting the CEFR levels of unseen learner texts than classification. Therefore, it is more appropriate to make use of the more informative ranking scores produced by the ranking model to learn a function that bands the scores into CEFR levels.
		
		In learning the mapping function, we adopted a five-fold cross-validation approach. We split the Cambridge Exams dataset into five cross validation folds, with approximately equal number of documents at each level in each fold. A mapping function that converts ranking scores into CEFR levels is learnt from training folds and then tested on the validation fold in each run. The final results are averaged across the runs.
		
		We compared three groups of methods to learn the mapping function. 
		
		{\bf (1) Regression and rounding}: A regression function is learnt from the ranking scores and the ground truth labels on the training part of the dataset and then applied to the validation part. The mapped CEFR prediction is then rounded to its closest integer and clamped to range $[1, 5]$. Both linear regression and polynomial regression models are considered. The intuition behind using polynomial functions instead of a simple linear function for mapping is that the correlation of ranking scores and CEFR levels is not necessarily linear so a non-linear function might be more suitable for this task.
		
		{\bf (2) Learning the cut-off boundary}: We learn a separation boundary that bands the ranking scores to levels by maximizing the accuracy of such separation. For instance, we consider the ranked documents as a list with descending readability, with their ranking scores following the same order. If we could find a suitable cut-off boundary between each two adjacent levels in the list, then every document above the boundary would fall into the higher level, and all documents below the boundary into the lower level. In this way, the ranked documents are banded into five levels with four separation boundaries learnt. 
		
		{\bf (3) Classification on the ranking scores}: The task can also be addressed as a classification problem. The ranking scores can be considered as a single dimensional feature and CEFR levels as the target value. Here, two approaches are adopted and compared, logistic regression and a linear SVM. As a matter of fact, the SVM approach can be considered as a variation of learning a separation boundary, as it tries to find an optimal decision boundary between the classes. 
		
		\begin{table}[t]
			\centering
			\small
			\begin{tabular}{l|c|c}
				\hline
				Mapping functions & ACC & PCC\\
				\hline \hline
				linear regression & 0.541 & 0.587\\
				\hline
				polynomial regression  & 0.586 & \textbf{0.873} \\
				\hline
				cut-off boundary & 0.562 & 0.872\\
				\hline
				logistic regression & 0.610 & 0.862\\
				\hline
				linear SVM & \textbf{0.622} & 0.864\\
				\hline
			\end{tabular}
			\caption{Results of mapping ranking scores to CEFR levels}
			\label{map}

		\end{table}
		
		Table \ref{map} shows the results of the three mapping methods.  Among the three approaches for mapping ranking scores to CEFR levels (regression-based, separation boundary-based, and classification-based), the classification ones showed better results than the others in terms of accuracy. Though not as high in accuracy as the SVM, a polynomial mapping function\footnote{
		A $4$th order polynomial function is adopted because it yields better results compared to other orders.} also yielded very good results in terms of $PCC$. Compared to the other two methods, the separation boundary-based approach performs better than a linear regression function but fails to match the polynomial regression and classification-based methods. Nonetheless, all three approaches considerably outperformed the naive generalization of the classification model from the WeeBit corpus to the Cambridge Exams data. These improvements are statistically significant at $p$$<$$0.05$ level.\footnote{Throughout this paper, we test  significance using $t$-test for $ACC$ and Williams' test \cite{williams1959regression} for $PCC$.}
		
\subsection{Domain Adaptation from Native to L2 Data}

    Another way to make use of the native data is to treat the task as a domain adaptation problem, where the WeeBit corpus is taken as the source domain, and the L2 data as the target domain. The idea behind this is to use out-of-domain training data to boost the performance on limited in-domain data. 
    
    EasyAdapt \cite{daume2007frustratingly} is one of the best performing domain adaptation algorithms. It has previously been applied to essay scoring and showed good results \cite{phandiflexible}. In a two domain case, EasyAdapt expands the input feature space from $\mathbb{R}^F$ to $\mathbb{R}^{3F}$, and then applies two mapping functions $\Phi^S(\textbf{x}) = \langle\textbf{x}, \textbf{x}, \textbf{0}\rangle$ and $\Phi^T(\textbf{x}) = \langle\textbf{x}, \textbf{0}, \textbf{x}\rangle$ on source domain data and target domain data input vectors respectively. Here, $\textbf{0} = \langle0,...0\rangle \in \mathbb{R}^F$ is the zero vector. In this manner, the instance feature vectors from the WeeBit corpus and Cambridge Exams datases are augmented to three times their original dimensionality. The augmented feature space captures both general and domain specific information and is thus capable of generalizing source domain knowledge to facilitate estimation on the target domain. As there is a mismatch between the levels on native and L2 data, the pairwise ranking algorithm needs to be adapted to ensure that the preference pairs are only created from the same domain. A five-fold cross-validation is used as in previous experiments.
    
    Table \ref{ea} shows the results of applying EasyAdapt with the ranking model. For comparison, we also present the results obtained when we apply the model trained on the native data to the L2 data directly, and the results obtained when we train the ranking model on the L2 data only. We can see that ranking with EasyAdapt outperforms the naive generalization approach significantly ($p$$<$$0.05$), but it does not beat the results obtained when training a model on L2 data directly.
    
    \begin{table}[t]
	\centering
	\small
	\begin{tabular}{l|c|c}
		\hline
		 & pairwise ACC & PCC\\
		\hline \hline
		EasyAdapt & 0.933 & 0.905\\ \hline
	    native data only &  0.913 & 0.880\\ \hline
		L2 data only & 0.943 & 0.913\\ \hline
	\end{tabular}
	\caption{Results of domain adaptation from native to language testing data}
	\label{ea}

    \end{table}

    After applying the ranking model with EasyAdapt, the ranking scores can be converted to CEFR levels using the same methods as described in Section \ref{sec:mapping}. The best mapped CEFR estimation is achieved with a linear SVM classifier on the ranking score, reaching an $ACC$ of $0.707$ and $PCC$ of $0.899$. Compared to the naive generalization of the classification model from native to L2 data, the mapped estimation is less influenced by the mismatch between difficulty levels in the two domains (see Table \ref{cm_easyadapt}).
    
        \begin{table}[t]
			\centering
			\small
			\resizebox{0.5\columnwidth}{!}{
			\begin{tabular}{l|ccccc}
				Levels & 1 & 2 & 3 & 4 & 5 \\
				\hline
				A2 & 11 & 3 & 0 & 0 & 0 \\
				B1 & 2 & 9 & 0 & 1 & 0 \\
				B2 & 0 & 0 & 13 & 0 & 2 \\
				C1 & 0 & 0 & 2 & 9 & 2\\
				C2 & 0 & 0 & 0 & 4 & 10 \\
			\end{tabular}}
			\caption{Confusion matrix of the mapped estimation after EasyAdapt application on one of the cross-validation folds}
			\label{cm_easyadapt}

		\end{table}

\subsection{Using Self-training to Enhance the Classification Model}

    In addition to the domain adaptation, we experimented with self-training to boost the performance on the limited L2 data with the native data. To the best of our knowledge, neither of the approaches has been applied to readability assessment before. 
    
    Self-training is a commonly used semi-supervised machine learning algorithm that aims to use the large amount of unlabelled data to help build a better classifier on a small amount of labeled data \cite{zhu2005semi}. When using native data to boost model performance on L2 data with self-training, the L2 data is regarded as labeled instances, and the native data as unlabeled ones. A model is trained on the L2 data and then used to score the native data. The most confident \textit{K} instances as well as their labels are added to the training set. Then the model is re-trained and the procedure is repeated. A five-fold cross-validation is used in evaluation as before.
    
    We have experimented with a grid search on \textit{K}'s and the number of iterations, and found out that whatever the choice of the parameters is, the model performance degrades with self-training when the unlabeled instances are added blindly to all levels of the L2 dataset. Taking into account the mismatch in the difficulty levels between the native and L2 texts, we adapted the algorithm to add the unlabeled data only to the lower three levels of the L2 dataset. The best result is achieved with $K$$=$$10$ and $9$ iterations, with $270$ texts added in total (as shown in Table \ref{st}). It seems reasonable to compare the results of this approach to those obtained with a model that is trained directly on the L2 data. Hence, we include the results of this model in Table \ref{st} for comparison.

    	\begin{table}[t]
	\centering
	\small
	\begin{tabular}{l|c|c}
		\hline
		Type & ACC & PCC\\
		\hline \hline
    	L2 data only & 0.785 & 0.924\\ \hline
		self-training & 0.797 & 0.938\\
		\hline
	\end{tabular}
	\caption{Results of self-training}
	\label{st}

    \end{table}

    The results show that self-training can significantly ($p$$<$$0.05$) help estimating readability for L2 texts by including a certain amount of unlabeled data (in this case, the native data) in training. However, the range of the reading difficulty covered by the unlabeled data may influence the model performance.

\section{Conclusions and Future Work}

    We investigated text readability assessment for both native and L2 learners. We collected a dataset with text tailored for language learners' readability and explored methods to adapt models trained on larger existing native corpora in estimating text reading difficulty for learners. In particular, we developed a system that achieves state-of-the-art performance in readability estimation, with $ACC$$=$$0.803$ and $PCC$$=$$0.900$ on native data, and $ACC$$=$$0.785$ and $PCC$$=$$0.924$ on L2 data, using a linear SVM. We compared a ranking model against the classification model for the task and showed that although a ranking model does not necessarily outperform a classification one in readability assessment on the same data, it is more accurate when generalizing to an unseen dataset. Following this, we showed that, by applying a ranking model and then learning a mapping function, the model trained on the native data can be applied to estimate the CEFR levels of unseen text effectively. This model achieves an accuracy of $0.622$ and $PCC$ of $0.864$, and considerably outperforms the naive generalization of the classification model, which achieves an accuracy of $0.233$ and $PCC$ of $0.730$. 
    
    In addition, we experimented with domain adaptation and self-training approaches to make use of the more plentiful native data to produce better estimation of readability when the L2 data is limited. When treating the native data as a source domain and L2 data as a target domain, applying the EasyAdapt algorithm for ranking achieves an accuracy of $0.707$ and  $PCC$$=$$0.899$. The best result is achieved by using self-training to include native data as unlabelled data in training the classification model, with $ACC$$=$$0.797$ and $PCC$$=$$0.938$.
    
    Future work will focus on the improvement of readability assessment framework for L2 learners and the identification of the optimal feature set that can generalize well to unseen text.

    \section*{Acknowledgements}
    We thank Cambridge Assessment for their assistance in the collection of the language testing data. We would like to express our gratitude to Sowmya Vajjala and Detmar Meurers for sharing the WeeBit corpus with us. We are also grateful to the reviewers for their useful comments. We thank Lucy Cavendish College for their support in the publication.

\bibliography{naaclhlt2016}
\bibliographystyle{naaclhlt2016}

\end{document}